\pdfoutput=1

\documentclass[11pt]{article}

\usepackage[final]{acl}

\usepackage{times}
\usepackage{latexsym}
\usepackage{multirow}
\usepackage{graphicx}
\usepackage{pdfpages}
\usepackage[T1]{fontenc}

\usepackage[utf8]{inputenc}

\usepackage{microtype}

\usepackage{inconsolata}
\usepackage{graphicx}
\usepackage{arydshln}
\usepackage{booktabs}
\usepackage{xcolor}
\usepackage{pifont}
\usepackage{soul}

\definecolor{myRed}{rgb}{0.808,0.067,0.149}
\definecolor{myGreen}{rgb}{0.067,0.708,0.149}

\newcommand{\xmark}{{\color{myRed}\ding{55}}}%
\newcommand{\cmark}{{\color{myGreen}\ding{51}}}

\title{A Novel Cartography-Based Curriculum Learning Method Applied on RoNLI: The First Romanian Natural Language Inference Corpus}

\author{\hspace{0.15cm}Eduard Poesina\\
  \hspace{0.15cm}University of Bucharest\\
  \hspace{0.15cm}Bucharest, Romania\\
  \hspace{0.15cm}\texttt{eduardgabriel.poe@gmail.com}\And
  \hspace{0.075cm}Cornelia Caragea\\
  \hspace{0.075cm}University of Illinois Chicago\\
  \hspace{0.075cm}Chicago, IL, USA\\
  \hspace{0.075cm}\texttt{cornelia@uic.edu}\And
  Radu Tudor Ionescu\thanks{$\;\;$Corresponding author.}\\
University of Bucharest\\
  Bucharest, Romania\\
  \texttt{raducu.ionescu@gmail.com}}

\begin{document}

\maketitle

\begin{abstract}
Natural language inference (NLI), the task of recognizing the entailment relationship in sentence pairs, is an actively studied topic serving as a proxy for natural language understanding. Despite the relevance of the task in building conversational agents and improving text classification, machine translation and other NLP tasks, to the best of our knowledge, there is no publicly available NLI corpus for the Romanian language. To this end, we introduce the first Romanian NLI corpus (RoNLI) comprising 58K training sentence pairs, which are obtained via distant supervision, and 6K validation and test sentence pairs, which are manually annotated with the correct labels. We conduct experiments with multiple machine learning methods based on distant learning, ranging from shallow models based on word embeddings to transformer-based neural networks, to establish a set of competitive baselines. Furthermore, we improve on the best model by employing a new curriculum learning strategy based on data cartography. Our dataset and code to reproduce the baselines are available at \url{https://github.com/Eduard6421/RONLI}.
\end{abstract}


\setlength{\abovedisplayskip}{3.0pt}
\setlength{\belowdisplayskip}{3.0pt}

\section{Introduction}

Given a sentence pair composed of a premise and a hypothesis, natural language inference (NLI) \cite{Korman-JASIST-2018}, a.k.a.~textual entailment recognition, is the task of determining if the premise entails, contradicts, or is neutral to the hypothesis. NLI is an actively studied problem \cite{bigoulaeva-etal-2022-effective,chakrabarty-etal-2021-figurative,chen-etal-2021-neurallog,luo-etal-2022-simple-challenging,mathur-etal-2022-docinfer,sadat-caragea-2022-scinli,sadat-caragea-2022-learning,snijders-etal-2023-investigating,varshney-etal-2022-unsupervised,wang-etal-2022-capture,wijnholds-2023-assessing}, being an essential task to be solved before addressing natural language understanding (NLU). Its complexity stems from the fact that NLU is generally considered an AI-hard problem \cite{Yampolskiy-AIECM-2013}. Notably, NLI forms the foundation of numerous advanced natural language processing systems \cite{Korman-JASIST-2018}, providing a backbone for multiple study areas such as language modeling \cite{merrill-etal-2022-entailment,mitchell-etal-2022-enhancing}, conversational agents \cite{raghu-etal-2022-structural}, zero-shot text classification \cite{pamies-etal-2023-weakly}, image captioning \cite{shi-etal-2021-enhancing}, text summarization \cite{falke-etal-2019-ranking}, 
discourse parsing \cite{sluyter-gathje-etal-2020-shallow}, and many others \cite{Korman-JASIST-2018}. The importance of NLI is well recognized, being included as a reference task in benchmarks such as GLUE \cite{Wang-ICLR-2019} and SuperGLUE \cite{Wang-NeurIPS-2019}.


\vspace{-2mm}
To date, the NLI task has been intensively studied in 
the English language \cite{bowman-etal-2015-large,koreeda-manning-2021-contractnli-dataset,romanov-shivade-2018-lessons,sadat-caragea-2022-scinli,N18-1101} and other languages, such as Chinese \cite{hu-etal-2020-ocnli}, Turkish \cite{budur-etal-2020-data}, Portugese \cite{Real-PROPOR-2020} and Indonesian \cite{mahendra-etal-2021-indonli}, as well as in multi-lingual scenarios \cite{conneau-etal-2018-xnli}. However, the sparsity of resources led researchers to overlook the development of NLI models for low-resource languages, such as Romanian \cite{Rotaru-NAACL-2024}. In this context, we introduce a novel \textbf{Ro}manian \textbf{N}atural \textbf{L}anguage \textbf{I}nference corpus (RoNLI) composed of 64K sentence pairs. The samples are divided into 58K for training, 3K for validation, and 3K for testing. We collected the sentence pairs from the Romanian version of Wikipedia, while searching for certain linking phrases between contiguous sentences to automatically label the sentence pairs. The training pairs are obtained via an automatic rule-based annotation process known as \emph{distant supervision} \cite{Mintz-ACL-2009}, while the validation and test sentence pairs are manually annotated with the correct labels. To the best of our knowledge, RoNLI is the first publicly available corpus for Romanian natural language inference. RoNLI is meant to attract the attention of researchers in studying and developing NLI models for under-studied languages, where NLU systems often produce subpar results, due to data scarcity and cross-lingual settings. To this end, we believe that RoNLI is a valuable resource. We publish our data and source code under the open-source CC BY-NC-SA 4.0 license at: \url{https://github.com/Eduard6421/RONLI}.

We carry out experiments with multiple machine learning methods, 
ranging from shallow models based on word embeddings to transformer-based neural networks, to establish a set of competitive baselines. Moreover, we employ data cartography \cite{swayamdipta-etal-2020-dataset} to characterize our training samples as easy-to-learn (E2L), ambiguous (A), and hard-to-learn (H2L), from the perspective of the best baseline model. Further, we study several approaches that harness the resulting data characterization to mitigate the inherent labeling noise caused by the automatic labeling process. We manage to improve the top model via a novel curriculum learning method based on data cartography.



\section{Related Work}







To date, there are multiple NLI datasets for the English language that are derived from {\em different data sources} and capture {\em a different label space}. For example, the {RTE} dataset \cite{dagan06rte}, which was crucial for the emergence of the NLI task, consists of premises extracted from news articles, with each \emph{<premise, hypothesis>} pair being labeled as either entailment or non-entailment. {SICK} \cite{marelli-etal-2014-sick} contains sentence pairs derived from image captions and video descriptions classified into three classes: entailment, contradiction, or neutral. Similar to SICK, {SNLI} \citep{bowman-etal-2015-large} consists of premises extracted from image captions (with hypotheses being written by human annotators), covering the same label space as SICK. Despite being very large in size, one limitation of {SNLI} is that it lacks data source diversity. \citet{N18-1101} introduced {MNLI} to specifically increase the diversity of the data. That is, in MNLI, the premises are extracted from diverse sources such as face-to-face conversations, travel guides, and the 9/11 event, while the hypotheses are derived exactly as in SNLI. 
While both SNLI and MNLI are widely used to track progress in natural language understanding, they have been shown to contain spurious correlations \cite{gururangan-etal-2018-annotation}. In an effort to reduce spurious correlations and create a harder dataset, \citet{nie-etal-2020-adversarial} developed {ANLI}, where human annotators wrote each hypothesis (conditioned on the premise and the label---entailment, contradiction, neutral) in an adversarial fashion, 
until the model failed to correctly predict the example. 

\vspace{-0.5mm}
There are many other relevant NLI datasets for English, which are briefly mentioned next. {QNLI} \cite{Wang-ICLR-2019} is derived from the SQuAD question answering dataset \cite{rajpurkar-etal-2016-squad}. {SciTaiL} \cite{khot2018scitail} is derived from a school level science question-answer corpus with the sentence pairs being classified into two classes: entailment or not-entailment. {MedNLI} \cite{romanov-shivade-2018-lessons} is obtained from medical records of patients with the \emph{<premise, hypothesis>} pairs being annotated by doctors as entailment, contradiction, or neutral. {ContractNLI} \cite{koreeda-manning-2021-contractnli-dataset} is derived from document-level contracts and annotated with a set of hypotheses (or claims) per contract with each \emph{<contract, hypothesis>} pair belonging to entailment, contradiction, or neutral. {SciNLI} \cite{sadat-caragea-2022-scinli} is obtained from research articles published in the ACL Anthology, but unlike the above NLI datasets where the hypotheses are written by humans specifically with the NLI task in mind, in SciNLI, both sentences in a pair are extracted from research articles. These sentences are written by researchers without the NLI task in mind, and thus, are expected to contain fewer spurious correlations. The labels in SciNLI are entailment, contrasting, reasoning, and neutral (extending the usual set with the reasoning relation that often occurs 
in formal writing). 

\begin{table}[t!]
    \centering
    \small
    \setlength\tabcolsep{0.18cm}
    \begin{tabular}{lll}
    \hline
    \textbf{Category} & \textbf{Romanian} & \textbf{English Translation}\\
    \hline
    \multirow{3}{*}{Contrastive} & {\color{blue}\^{I}n contrast} & In contrast \\ 
    & {\color{blue}\^{I}n contradic\c{t}ie} & In contradiction \\
    & {\color{blue}\^{I}n opozi\c{t}ie}  & In opposition \\
    \hline
    \multirow{3}{*}{Entailment} & {\color{blue}Cu alte cuvinte} & In other words \\     
    & {\color{blue}\^{I}n al\c{t}i termeni} & In other terms \\
    & {\color{blue}Pe larg} & In broader terms \\ 
    \hline
    \multirow{3}{*}{Reasoning} & {\color{blue}Astfel} & Thus \\
    & {\color{blue}Prin urmare} & Therefore \\
    & {\color{blue}\^{I}n concluzie} & In conclusion \\
    \hline
    Neutral & N/A & N/A \\
    \hline
    \end{tabular}
    \vspace{-2mm}
\caption{Examples of original (Romanian) and translated (English) linking phrases and transition words that suggest certain relations between two sentences. Upon finding a premise and a hypothesis linked by one such phrase, we remove the respective phrase to force NLI models trained on the extracted sentences to focus on other clues.\vspace{-2mm}}
\label{tab:link_phrases}
\end{table}

\begin{table*}[t]
    \centering
    \small
    \begin{tabular}{lcccccccccccc}
    \hline
    \multirow{2}{*}{\textbf{Relation}} & \multicolumn{3}{c}{\multirow{2}{*}{\textbf{\#Samples}}} & \multicolumn{6}{c}{\textbf{Average \#Words}} & \multicolumn{3}{c}{\multirow{2}{*}{\textbf{Overlap Ratio}}} \\
     & & & & \multicolumn{3}{c}{\textbf{Premise}} & \multicolumn{3}{c}{\textbf{Hypothesis}} & \\
    \hline
    & Train & Val & Test & Train & Val & Test & Train & Val & Test & Train & Val & Test      \\
    \cline{2-13}
    Contrastive & 2,592 & 74 & 74 & 26.7 & 25.7 & 24.6 & 25.4 & 27.1 & 23.5 & 0.07 &  0.07 & 0.08\\
    Entailment  & 1,300 & 72 & 96 & 26.6 & 25.6 & 24.5 & 23.4 & 20.9 & 23.0 & 0.07 & 0.08 & 0.07 \\
    Causal      & 25,722 & 1,134 & 952 & 25.0 & 25.3 & 25.4 & 23.7  & 23.8 & 23.3 & 0.06 & 0.05 & 0.05 \\
    Neutral   & 28,500 & 1,778 & 1,878 & 23.8 & 23.5 & 23.8 & 23.8 & 23.9 & 24.3 & 0.03 & 0.03 & 0.04\\
    \hline
    Overall & 58,114 & 3,059 & 3,000 & 24.5 &  24.4 & 24.3 & 23.7 & 23.9 & 23.8 & 0.04 & 0.04 & 0.04 \\
    \hline
    \end{tabular}
    \vspace{-2mm}
\caption{Statistics for each of the training, validation and test splits of the RoNLI dataset, including the class distribution (2nd to 4th columns), the average number of words in a premise (5th to 7th columns), the average number of words in a hypothesis (8th to 10th columns), and the average ratio of overlapping words between a premise and a hypothesis (11th to 13th columns).\vspace{-2mm}}
\label{tab:stats}
\end{table*}

\vspace{-0.5mm}
Owing to the importance of the NLI task, several efforts have been directed to the development of datasets in other languages. For example, 
\citet{conneau-etal-2018-xnli} introduced {XNLI}, a cross-lingual evaluation dataset obtained by translating examples from {MNLI}. \citet{hu-etal-2020-ocnli} created OCNLI, the first large Chinese dataset for NLI, whereas \citet{kovatchev-taule-2022-inferes} explored NLI in the Spanish language. 
Other works focused on the creation of NLI datasets from low-resource languages such as Creole \cite{armstrong-etal-2022-jampatoisnli}, Indonesian \cite{mahendra-etal-2021-indonli}, and Turkish \cite{budur-etal-2020-data}, either by human annotations or automatic translations from English. Similar to these latter works, we focus on a low-resource language, Romanian, to create a large NLI dataset, the first of its kind. In contrast to these studies and inspired by the creation of SciNLI \cite{sadat-caragea-2022-scinli}, instead of resorting to English translations, we use the linking phrases between contiguous sentences in text to automatically label a large training set for the Romanian language, while the validation and test sets are manually annotated. 

\section{Corpus}

\paragraph{Data gathering and automatic labeling.}
The data source for building our corpus is the Romanian Wikipedia, chosen for its sizable, diverse and expansive content generated by a wide user base. Wikipedia presents the particularity of allowing users to define special pages: help pages, discussion pages, disambiguation pages, and pages dedicated to hosting meta-information about uploaded files. We consider that such pages are less relevant to the NLI task, and hence, we decided to filter them out.
Next, we develop and apply a custom text parser to systematically deconstruct the corpus into individual articles, sub-articles, and chapters. The resulting text samples are further split using the sentence tokenizer available in NLTK \cite{BirdKleinLoper09}. 
However, not all sections of Wikipedia articles are equally valid sources to extract sentence pairs for natural language inference. To ensure the quality and relevance of our dataset, we exclude segments such as the list of references, the image XML markers, the external links to other articles, and the metadata articles, which usually contain unstructured or less meaningful text.

Due to the collaborative nature of Wikipedia's content, there is a high variety in the quality of sentences. We observed that shorter sentences tend to introduce content that is often deemed unrepresentative for the NLI task, lowering the overall quality of our dataset. 
To extract high-quality sentences from Wikipedia, we thus set a minimum length of 50 characters for each sentence. We select this limit empirically, based on visually inspecting the sentences extracted with our parsing methodology.

Building upon the methodology presented by \citet{sadat-caragea-2022-scinli}, we divide the sentence pairs into four separate relationship types: contrastive, entailment, reasoning and neutral. To automatically label the sentence pairs using distant supervision, we establish a set of language-specific linking phrases and transition words that are very likely to predetermine the relationship between two sentences. Some examples of linking phrases for the related sentence pairs are shown in Table \ref{tab:link_phrases}. Note that, for neutral sentence pairs, we require the absence of linking phrases. The full set of linking phrases for each relationship type (contrastive, entailment, reasoning) are listed in Appendix \ref{sec:appendix}. For instance, starting a sentence with the phrase ``\^{I}n concluzie'' (translated to ``In conclusion'') could be easily identified as a predictive cue, suggesting a reasoning relationship with the preceding sentence. 

While recognizing that this approach is not infallible, it is undoubtedly an effective strategy for automatically collecting representative data for NLI in an unsupervised manner. The method allows us to harness the inherent relational information embedded in the Romanian vocabulary without requiring specific knowledge about the subject matter. By employing this methodology, we successfully extract a total of 64K sentence pairs. In all the extracted examples, the linking phrases are removed to make sure that models do not learn superficial patterns (based on linking phrases) from the data. 

To create our training, validation, and test sets, we employ stratified sampling to maintain the inherent distribution of the original data. In Table \ref{tab:stats}, we report several statistics for each of the three subsets, such as the class distribution, the average number of words in a premise, the average number of words in a hypothesis, and the average ratio of overlapping words between a premise and a hypothesis. Notably, we observe that the chosen data source, Wikipedia, leads to an uneven sample distribution, with significantly fewer sentences being attributed to the contrastive and entailment classes. 




\vspace{-2.8mm}
\paragraph{Manual labeling.}
To ensure that the validation and test 
labels are reliable, and models do not reach good performance levels due to biases in our automatic annotation process, we instruct three human annotators who are native Romanian speakers (with bachelor degrees) to manually relabel our validation and test sentences. To avoid biasing the annotators towards preferring the automatic labels, we preclude them from viewing the automatically generated labels (and the linking phrases). Our annotators received detailed guidelines and examples to ensure a consistent annotation process. The inter-rater Fleiss Kappa coefficient is 0.71, which denotes a substantial agreement between annotators. Notably, the agreement level among our annotators is larger than the Kappa agreement of 0.62 reported for SciNLI \cite{sadat-caragea-2022-scinli}, and the Kappa agreement of 0.70 reported for SNLI \cite{bowman-etal-2015-large}. We originally allocated more data samples to the validation and test sets, but discarded validation and test samples that had no majority label provided by the annotators. The reported numbers of data samples for validation and test do not include the discarded ones. We compared the automatic annotations with the aggregated manual labels, obtaining a Cohen's Kappa coefficient between the automatic and manual labels of 0.62, which indicates a substantial agreement. Based on this high agreement, we have strong reasons to believe that the automatic labeling process is sufficiently accurate.


\begin{figure*}[!t]
\begin{center}
\centerline{\includegraphics[width=0.98\linewidth]{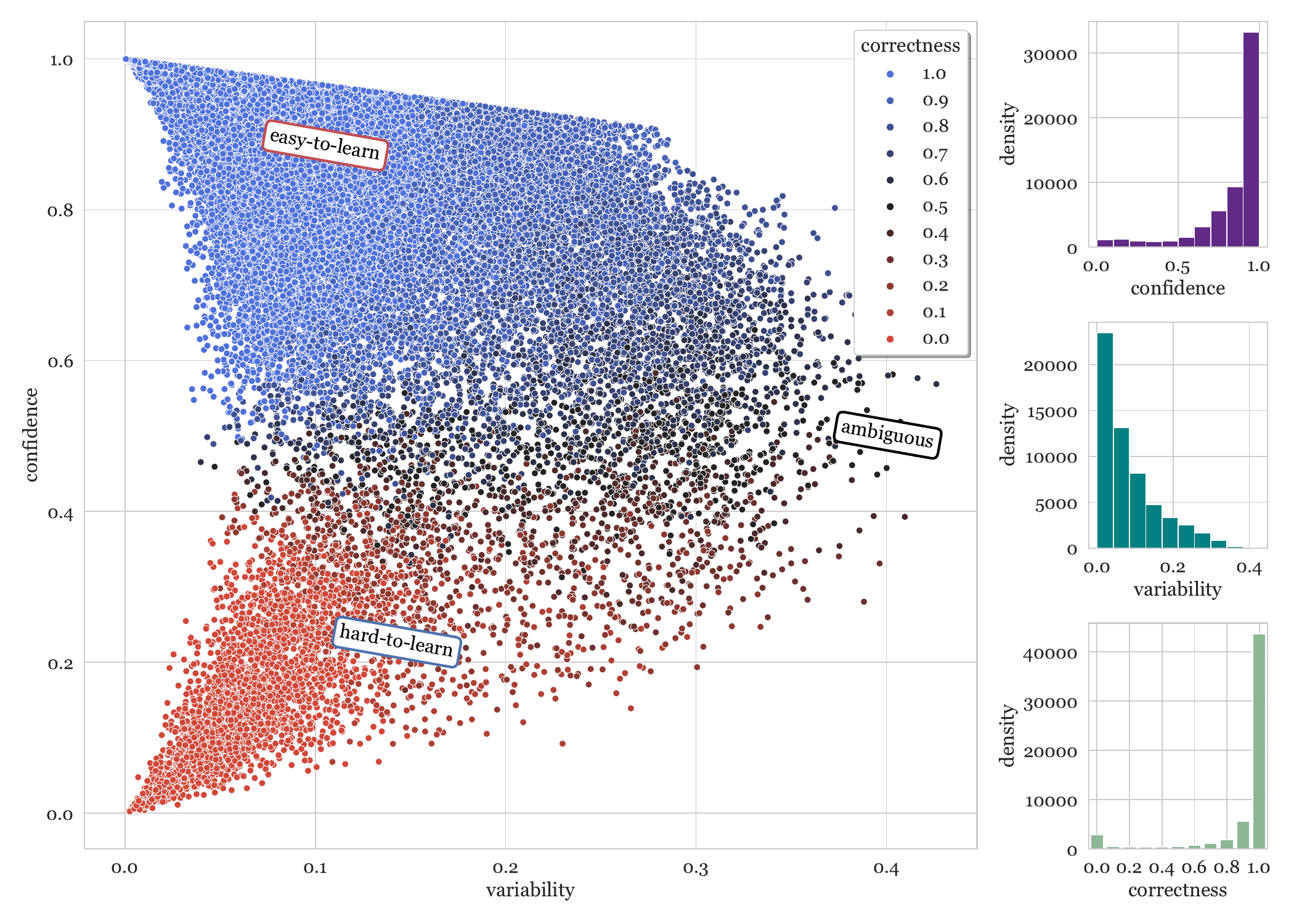}}
\vspace{-0.4cm}
\caption{Data cartography visualization of the RoNLI dataset based on fine-tuning the Ro-BERT model \cite{Dumitrescu-EMNLP-2020}. In the left-hand side plot, the $y$-axis corresponds to the level of confidence exhibited by the model during training, while the $x$-axis represents the variability of the confidence level. Adjacent to the primary plot, three histograms are displayed on the right-hand side, each representing a different metric: the confidence, the variability of confidence, and the correctness. The visualization offers a comprehensive overview of our dataset characteristics and the behavior of Ro-BERT during training. Best viewed in color.\vspace{-2mm}}
\label{fig:cartography}
\vspace{-0.6cm}
\end{center}
\end{figure*}


\vspace{-2.8mm}
\paragraph{Data cartography.} To further investigate the correctness and utility of the collected samples, we employ data cartography \cite{swayamdipta-etal-2020-dataset} on top of our best model, a version of the Romanian BERT (Ro-BERT) \cite{Dumitrescu-EMNLP-2020} fine-tuned on the RoNLI training data. \citet{swayamdipta-etal-2020-dataset} proposed \emph{data cartography} as a visualization technique based on interpreting the behavior of a machine learning model during the training stage. The method is built upon two insightful metrics recorded during the training process, specifically the level of \emph{confidence} when accurately categorizing a data point into the correct class, and the \emph{variability} (fluctuation) of the confidence during training. They are complemented by \emph{correctness}, a metric that tells how often a sample is correctly classified during training.

As illustrated in Figure \ref{fig:cartography}, data cartography provides a characterization of the training examples as easy-to-learn (E2L), ambiguous (A), and hard-to-learn (H2L), for the chosen Ro-BERT model. 
Easy-to-learn examples reside in the top-left quadrant of the data map, describing items which exhibit low variability and high confidence. The model's strong certainty in its decisions indicates that it faces minimal difficulty in comprehending these examples.
In contrast, hard-to-learn examples, occupying the bottom-left quadrant, demonstrate low variability but differentiate themselves from the previous category in terms of confidence, which stays low in this case. 
The low confidence of the model in the ground-truth label suggests that the examples from this region are likely to be mislabeled. This is because most examples from the underrepresented classes (contrastive, entailment) are labeled as H2L (see Table \ref{tab:carto_distrib}).
In contrast, ambiguous examples are characterized by high variability and mid-level confidence, and are thus positioned on the middle-right side of the map. Frequent changes in the predicted classes from one epoch to another prove that these examples are very challenging for the model.  

For most training examples, our data map shows that Ro-BERT presents predominantly high confidence (over 30K training samples have the average confidence above 0.9) and high correctness (over 40K samples are correctly classified), suggesting that the model is particularly effective in recognizing and learning consistent patterns. A secondary cluster can be observed pertaining to the hard-to-learn scenarios reflecting that the model starts and remains strongly biased across training epochs for a relatively small portion of samples. In summary, the data map suggests that most samples are consistently labeled. Corroborating this observation with the substantial Cohen's Kappa agreement between the automatic and manual labels suggests that the automatic labels are mostly correct. Nonetheless, we further investigate the implications of the observed data distribution in Section \ref{sec_results} by considering training scenarios where we deliberately choose the type of data to incorporate, encompassing varying combinations of E2L, A, and H2L instances. 



\section{Models}
\label{sec:baselines}

We employ grid-search to tune the hyperparameters of all models (see details in Appendix \ref{sec:appendix}). 

\vspace{-2.6mm}
\paragraph{Classifiers based on word embeddings.} 
Our first two baselines are based on word embeddings, leveraging the 300-dimensional word vectors returned by the fastText framework \cite{bojanowski2017enriching}, which includes support for the Romanian language. We select fastText primarily due to its inherent capability to manage out-of-vocabulary words, a crucial aspect considering the diverse nature of Wikipedia content. 
We represent each sentence as a continuous bag-of-words (CBOW), i.e., we compute the average of the word embeddings in each sentence. Next, the CBOW representations generated for a sentence pair are concatenated, resulting in a joint representation of the premise and the hypothesis. We consider two alternative classifiers to learn from the extracted CBOW embeddings, namely a linear Support Vector Machines (SVM) model and a Multinomial Logistic Regression (Softmax) model.

\vspace{-2.6mm}
\paragraph{RoGPT2.} 
The next baselines draw upon the capabilities of transformer models \cite{Vaswani-NIPS-2017}, which have shown impressive performance on various NLP tasks, including NLI. Our third baseline is based on the generative capabilities of the GPT family of models. By harnessing the inherent generative knowledge embedded within GPT, we can establish meaningful correlations between two sentences, a premise and a hypothesis, and accurately infer their relationship. For this approach, we employ a GPT2 model pre-trained on Romanian corpora \cite{niculescu2021rogpt2}. We extract the last hidden states of the end-of-sentence (EOS) token and feed it to a softmax classification head.

\vspace{-2.6mm}
\paragraph{ChatGPT 3.5 Turbo.} Another way to utilize transformer-based large language models (LLMs) for various tasks is via in-context (prompt-based) learning. This approach leverages the meta-learning capabilities of LLMs. To this end, we employ a powerful multilingual LLM, namely ChatGPT 3.5 Turbo \cite{Ouyang-NeurIPS-2022}, with in-context learning (two examples per class), to directly predict the test labels. This model does not require any fine-tuning.




\vspace{-2.6mm}
\paragraph{Multilingual BERT.} 
We employ a multilingual BERT model which is pre-trained on the XNLI corpus \cite{conneau-etal-2018-xnli}. We then consider two options, one based on zero-shot learning, and one based on fine-tuning. The zero-shot multilingual BERT follows the Definition-Restrictive approach described by \citet{yin-etal-2019-benchmarking}. This is because our set of classes (defined in Appendix \ref{sec:appendix}) do not coincide exactly with those of XNLI.

\vspace{-2.6mm}
\paragraph{Ro-BERT.} 
Our sixth baseline is a variant of BERT \cite{Devlin-NAACL-2019} pre-trained on Romanian corpora, namely Ro-BERT \cite{Dumitrescu-EMNLP-2020}. We hypothesize that Ro-BERT discerns inherent characteristics within and between sentences, allowing it to accurately predict the relationship type. We extract the [CLS] token from the final hidden state of each sentence in order to obtain an internal representation. The extracted embeddings are then concatenated and fed into a softmax classification head, enabling the system to predict the relationship between pairs.



\vspace{-2.6mm}
\paragraph{Ro-BERT (spurious).}
Natural language inference datasets commonly face the challenge of spurious correlations \cite{gururangan-etal-2018-annotation}, which are statistical associations that machine learning models learn to interpret and rely upon when making predictions. Such correlations do not necessarily reflect a robust understanding of NLI, but rather illustrate common language patterns and expressions. In order to assess how much RoNLI is affected by this phenomenon, we fine-tune Ro-BERT only on the hypotheses, with the same hyperparameters as above. The performance of this model is proportional to the amount of spurious correlations.

\vspace{-2.6mm}
\paragraph{Ro-BERT + cartography subsampling.}
We explore additional baselines based on the categorization produced by data cartography. Following the work of \citet{swayamdipta-etal-2020-dataset}, we fine-tune the Ro-BERT model on subsets of the training set, determined by the difficulty categories: easy-to-learn, ambiguous and hard-to-learn. We train one Ro-BERT model for each of the three categories of samples. In addition, we train a Ro-BERT model on easy-to-learn and ambiguous (E2L+A) samples, based on the intuition that hard-to-learn samples suffer from high rates of labeling noise, and should thus be discarded.

\vspace{-2.6mm}
\paragraph{Ro-BERT + Length-CL.}
We employ a Ro-BERT model based on curriculum learning, a training paradigm proposed by \citet{Bengio-ICML-2009}. The technique emulates the human learning process, which typically progresses from simpler to more complex concepts, and applies it to the training of machine learning models. The idea behind the technique is to expose the model to easier examples first, thus building a foundational understanding of the task, which will in turn allow it to accommodate more complex samples faster. Curriculum learning has received significant attention from researchers due to its vast applicability, as shown by \citet{Soviany-IJCV-2022}. Following \citet{Nagatsuka-NGC-2023}, the considered baseline applies the curriculum with respect to text length.

For a fair comparison with the conventional training paradigm, we ensure that the total number of training iterations is equal with those of the base model. We perform the curriculum training for half the number of total iterations. Subsequently, we continue training under the standard paradigm, using all available data for the remaining iterations. As for the baseline Ro-BERT, we introduce early stopping to prevent overfitting. We apply the same training procedure to all curriculum models described below.

\vspace{-2.6mm}
\paragraph{Ro-BERT + STS-CL.}
An intuitive way to introduce curriculum learning with respect to the NLI task is to use the semantic text similarity (STS) between sentences to measure difficulty. We thus add a baseline Ro-BERT model that uses curriculum learning based on STS. To calculate STS, we utilize the cosine similarity based on the fine-tuned Ro-BERT embeddings of the sentences. The training starts with the most similar sentence pairs, and progressively adds less similar sentence pairs. 

\vspace{-2.6mm}
\paragraph{Ro-BERT + Cart-CL.}
We propose a Ro-BERT model based on curriculum learning with respect to the difficulty groups established via data cartography. We start the training on the E2L samples for a number of iterations. We gradually incorporate more challenging batches, adding ambiguous data instances, then H2L samples. If the total number of iterations is $N$, we perform $N/4$ iterations with E2L samples, $N/4$ iterations with A samples, and $N/2$ iterations with H2L samples. 


\begin{table}[t]
\centering
\small
\setlength\tabcolsep{0.15cm}
\begin{tabular}{ccccc}
\hline
& \textbf{Contrastive} & \textbf{Entailment} & \textbf{Causal} & \textbf{Neutral} \\
\hline
E2L & 0 & 0 & 2,207 & 17,082 \\
A & 967 & 375 & 9,614 & 8,334 \\
H2L & 2,592 & 1,300 & 11,632 & 3,765\\
\hline
\end{tabular}
\vspace{-2mm}
\caption{Distribution of classes in each difficulty group defined via data cartography.}
\vspace{-2mm}
\label{tab:carto_distrib}
\end{table}

\vspace{-2.6mm}
\paragraph{Ro-BERT + Cart-CL++.}
Our next model confronts certain caveats introduced by data cartography \cite{swayamdipta-etal-2020-dataset} in terms of data difficulty characterization. Given that the grouping of data instances relies on one metric (either confidence or variability) instead of both, we observed that the E2L, A, and H2L groups defined by \citet{swayamdipta-etal-2020-dataset} do not form a partition of the training set, leaving samples outside these groups, while assigning other samples to two different groups. Leaving data outside or duplicating some of the examples can bias the model and lead to suboptimal performance. 
We conjecture that a better curriculum can be created by designing a difficulty scoring function that jointly takes confidence and variability into account. Let $c_i$ and $v_i$ be the confidence and variability of the $i$-th training sample. We introduce a new difficulty scoring function $s:[0,1]\times[0,1]\rightarrow[0,3]$ defined as follows:
\begin{equation}
\label{eq_multilabel}
    s(c_i,v_i) = \left\{
    \begin{array}{ll}
     1-c_i + v_i, \textnormal{if}\; c_i > 0.5 \\
     3-c_i - v_i, \textnormal{otherwise}
\end{array} ,
    \right.
\end{equation}
where $i \in \left\{1, ..., n\right\}$, and $n$ is the number of training samples. By design, our novel scoring function $s$ assigns low scores to items characterized by low variability and high correctness, medium scores for items perceived as ambiguous, and high scores to difficult examples.

\begin{table*}[t]
\centering
\small
\setlength\tabcolsep{0.15cm}
\begin{tabular}{lccccccccccccc}
\hline
\multirow{2}{*}{\textbf{Method}} & \textbf{Over} &  \multicolumn{3}{c}{\textbf{Contrastive}} & \multicolumn{3}{c}{\textbf{Entailment}} & \multicolumn{3}{c}{\textbf{Reasoning}} & \multicolumn{3}{c}{\textbf{Neutral}}\\
& \textbf{sampling} &  $\mathbf{P}$ & $\mathbf{R}$ & $\mathbf{F_1}$ &  $\mathbf{P}$ & $\mathbf{R}$ & $\mathbf{F_1}$ &  $\mathbf{P}$ & $\mathbf{R}$ & $\mathbf{F_1}$ &  $\mathbf{P}$ & $\mathbf{R}$ & $\mathbf{F_1}$\\
\hline
CBOW + SVM  & \xmark & 0.16 & {0.32} & 0.22 & 0.19 & 0.51 & {0.28} & 0.51 & 0.67 & 0.58 & 0.88 & 0.63 & 0.74 \\
CBOW + Softmax & \xmark & {\color{blue}\bf 0.50} & {0.01} & 0.03 & {\color{blue}\bf 0.57} & 0.04 & {0.08} & 0.49 & 0.81 & 0.61 & 0.85 & 0.65 & 0.74 \\
\hline
RoGPT2 & \xmark& 0.00 & 0.00 & 0.00  & 0.00 & 0.00 & 0.00  & 0.52 & {0.83} & 0.63 & 0.87 & 0.69 & 0.77 \\
ChatGPT 3.5 Turbo & - & 0.13 & 0.20 & 0.16 & 0.14 & 0.55 & 0.22 & 0.45 & 0.72 & 0.56 & {\color{blue}\underline{0.96}} & 0.51 & 0.66 \\
Zero-shot multilingual BERT & \xmark & 0.02 & {0.77} & 0.05  & 0.03 & 0.23 & 0.05  & 0.37 & {0.15} & 0.21 & 0.49 & 0.01 & 0.02 \\
Fine-tuned multilingual BERT & \xmark & 0.00 & 0.00 & 0.00  & 0.00 & 0.00 & 0.00  & 0.38 & {0.89} & 0.53 & 0.85 & 0.35 & 0.49 \\
Ro-BERT & \xmark & 0.04 & 0.01 & 0.01  & 0.00 & 0.00 & 0.00  & 0.55 & 0.86 & {0.67} & {0.89} & {0.72} & {0.79} \\
Ro-BERT (spurious) & \xmark & 0.07 & 0.01 & 0.03 & 0.00 & 0.00 & 0.00 & 0.53 & 0.69 & 0.60 & 0.80 & {\color{blue}\underline{0.75}} & 0.77 \\
\hline
Ro-BERT & \cmark & 0.19 & {\color{blue}\underline{0.80}} & 0.31 & {0.35} & 0.54 & {\color{blue}\underline{0.41}} & 0.62 & 0.72 & 0.67 & {\color{blue}\underline{0.96}} & 0.73 & {\color{blue}\underline{0.83}} \\
\hline
Ro-BERT + E2L & \cmark & 0.00 & 0.00 & 0.00  & 0.00 & 0.00 & 0.00  & {0.52} & 0.49 & 0.50 & 0.71 & {\color{blue}\bf 0.79} & 0.74 \\
Ro-BERT + A   & \cmark & 0.17 & 0.07 &  0.10  & {0.31} &  0.06 & 0.10  & 0.55 & {\color{blue}\underline{0.95}} & {\color{blue}\bf{0.70}} &  {\color{blue}\bf{0.97}} & 0.68 &  0.80 \\
Ro-BERT + H2L & \cmark & 0.00 & 0.00 & 0.00  & 0.00 & 0.00 & 0.00  & 0.31 &  {\color{blue}\bf{0.99}} & 0.48 &  0.46 & 0.00 & 0.00 \\
Ro-BERT + E2L + A & \cmark & 0.21 & 0.10 & 0.13  & 0.35 & 0.07 & 0.11 & 0.55 & 0.93 & {\color{blue}\underline{0.69}} & {\color{blue}\underline{0.96}} & 0.68 & 0.80 \\
\hline
{Ro-BERT + Length-CL} & \cmark & 0.17 & 0.72 & 0.28  & 0.34 & 0.43 & 0.36  & {0.62} & 0.71 & 0.66 & 0.95 & {\color{blue}\underline{0.75}} & {\color{blue}\bf{0.84}} \\
{Ro-BERT + STS-CL} & \cmark & 0.20 & {\color{blue}\bf{0.86}} &	{\color{blue}\underline{0.32}} & 0.37	& 0.40 & 0.37 & {\color{blue}\underline{0.63}} & 0.76 & {\color{blue}\underline{0.69}} & {\color{blue}\underline{0.96}} & 0.74 & {\color{blue}\bf{0.84}} \\
Ro-BERT + Cart-CL & \cmark & 0.13 & 0.63 & 0.21 &  0.22  & {\color{blue}\bf 0.62}  & 0.32 & {\color{blue}\bf{0.64}} & 0.63 & 0.63 & {\color{blue}\underline{0.96}} & {0.74} & {\color{blue}\bf{0.84}} \\
Ro-BERT + Cart-CL++ & \cmark & 0.19 & {0.79} & {0.31}  & {0.36} & {\color{blue}\underline{0.58}} & {\color{blue}\bf 0.44}  & {\color{blue}\underline{0.63}} & 0.74 & 0.68 & {\color{blue}\underline{0.96}} & {0.74} & {\color{blue}\bf{0.84}} \\
Ro-BERT + Cart-Stra-CL++ & \cmark & {\color{blue}\underline {0.26}} & {0.77} & {\color{blue}\bf 0.38}  & {\color{blue}\underline{0.45}} & 0.44 &  {\color{blue}\bf{0.44}} & 0.62 & {0.79} & {\color{blue}\bf 0.70} & {\color{blue}\underline{0.96}} & {0.74} & {\color{blue}\bf 0.84} \\
\hline
\end{tabular}
\vspace{-2mm}
\caption{Per class precision, recall and $F_1$ scores of the proposed baseline models on the manually labeled RoNLI test set. The scores are independently reported for each class to enable a detailed class-based assessment of the classification performance. The best results are shown in bold blue and the second best are underlined and in blue.\vspace{-2mm}}
\label{tab:results}
\end{table*}

\begin{table}[t]
\centering
\small
\setlength\tabcolsep{0.1cm}
\begin{tabular}{lccc}
\hline
\multirow{2}{*}{\textbf{Method}} & \textbf{Over} & \multicolumn{2}{c}{$\mathbf{F_1}$} \\
& \textbf{sampling} & \textbf{Micro} &  \textbf{Macro} \\
\hline
CBOW + SVM & \xmark& 0.63 & 0.45 \\
CBOW + Softmax & \xmark& 0.66 & 0.36 \\
\hline
RoGPT2 & \xmark& 0.70 & 0.30 \\
ChatGPT 3.5 Turbo & - & 0.57 & 0.40 \\
Zero-shot multilingual BERT & \xmark& 0.08 & 0.08 \\
Fine-tuned multilingual BERT & \xmark& 0.50 & 0.30 \\
Ro-BERT &\xmark & 0.72 & 0.37 \\
Ro-BERT (spurious) & \xmark & 0.69 & 0.35 \\
\hline
Ro-BERT & \cmark & {0.73} & 0.56 \\
\hline
Ro-BERT + E2L & \cmark & 0.65 & 0.31 \\
Ro-BERT + A & \cmark &  {0.73} & 0.44 \\
Ro-BERT + H2L & \cmark & 0.31 &  0.12 \\
Ro-BERT + E2L + A & \cmark & {0.73} & 0.44 \\
\hline
{Ro-BERT + Length-CL} & \cmark & {0.73} & 0.54 \\
{Ro-BERT + STS-CL} & \cmark & {\color{blue}\underline{0.74}} & {\color{blue}\underline{0.57}} \\

Ro-BERT + Cart-CL & \cmark & 0.70 & 0.51 \\
Ro-BERT + Cart-CL++ & \cmark & {0.73} & {\color{blue} \underline{0.57}} \\
Ro-BERT + Cart-Stra-CL++ & \cmark & {\color{blue} \bf 0.75} & {\color{blue} \bf 0.59} \\
\hline
\end{tabular}
\vspace{-2mm}
\caption{Overall micro and macro $F_1$ scores of the proposed baseline models on the manually labeled RoNLI test set. The micro and macro $F_1$ scores are both reported to acknowledge the behavior of models on our imbalanced dataset. The best results are shown in bold blue and the second best are underlined and in blue.}
\label{tab:overallresults2}
\vspace{-2mm}
\end{table}

\vspace{-2.6mm}
\paragraph{Ro-BERT + Cart-Stra-CL++.}
By analyzing the distribution of classes in each data cartography group (see Table \ref{tab:carto_distrib}), we observe that the E2L group does not contain any contrastive or entailment instances. Hence, we propose yet another model based on curriculum learning, which constructs stratified easy-to-hard batches. This ensures the diversity of class labels right from the beginning of the training process, avoiding to bias the model towards certain classes. This novel curriculum learning approach is more suitable for imbalanced datasets, which are more prone to be affected by introducing further class biases.

\section{Results and Observations}
\label{sec_results}


\paragraph{Experimental setup.}
We perform NLI experiments with the set of baselines described above. We also carry out experiments with the data cartography groups, considering scenarios when our top performing model is trained with subsets of the training set, namely using only E2L examples, A examples, H2L examples, and E2L+A examples. We add the models based on curriculum learning to our line-up of models, specifically Length-CL, STS-CL, Cart-CL, Cart-CL++, and Cart-Stra-CL++. As evaluation metrics, we report the precision, recall and $F_1$ score for each of the four classes, on the manually labeled test set. We also report the micro and macro $F_1$ scores to determine how models are affected by the imbalanced nature of RoNLI. Note that the micro $F_1$ is equivalent to the classification accuracy, while the macro $F_1$ is also known as the group-averaged $F_1$.

Our experimental methodology is based on conducting multiple trials of each experiment, specifically five runs, to ensure the robustness and reliability of our findings. The model that demonstrates the best performance on the validation set across the five trials is the one selected for evaluation on the test set. This procedure is fairly performed for all models, including the baselines.

\vspace{-2.6mm}
\paragraph{Results.}
The results obtained by the baselines on the individual classes are presented in Table \ref{tab:results}. The zero-shot and fine-tuned versions of multilingual BERT obtain subpar results, suggesting that cross-lingual systems are not suited for Romanian NLI. Among the various language models, Ro-BERT achieves the highest performance level. Still, we observe that models based on word embeddings are more able to learn the underrepresented classes than the vanilla transformers (despite the strong performance of transformer models on NLI for the English language). To overcome this limitation, we retrain the most promising transformer (Ro-BERT) by oversampling the underrepresented classes such that the class distribution becomes balanced. Since the results indicate that oversampling brings significant performance gains, we integrate oversampling in the training process of all the subsequent versions of Ro-BERT.


The overall results in terms of the micro and macro $F_1$ scores obtained by the baselines are shown in Table \ref{tab:overallresults2}. We observe that no baseline is capable of surpassing a micro-averaged $F_1$ score of 80\% and a macro-averaged $F_1$ score of 60\%. This observation highlights the complexity of the RoNLI dataset, pointing to a new challenging benchmark for natural language inference. 

The various training scenarios applied on Ro-BERT lead to some interesting observations. When Ro-BERT is trained on hypotheses, its performance drops by $3\%$ in terms of micro $F_1$, and $2\%$ in terms of macro $F_1$, indicating that spurious correlations are not enough to reach high performance levels. 

The Ro-BERT trained on E2L examples performs worse than the vanilla Ro-BERT, suggesting that the former model does not benefit much from the homogeneity of the easy examples. The Ro-BERT trained on H2L examples reaches an even lower performance, suggesting that focusing too much on complex examples can hurt overall performance, possibly due to overfitting on these hard cases (and their potentially erroneous labels). In contrast, the Ro-BERT trained on ambiguous examples and the one trained on easy and ambiguous examples (E2L + A) are fairly close to vanilla Ro-BERT, indicating that the ambiguous samples represent the most useful group to train the model on.




The state-of-the-art curriculum method, Length-CL \cite{Nagatsuka-NGC-2023}, as well as
our first curriculum learning approach, Cart-CL, exhibit lower performance than the vanilla Ro-BERT. Cart-CL++ outperforms Cart-CL, confirming that our difficulty scoring function $s$ is useful. Furthermore, our function boosts the macro $F_1$ score of Ro-BERT + Cart-CL++ over the vanilla Ro-BERT. However, it is likely that Cart-CL++ is still suboptimal because, in the early stages of the curriculum, when using only the E2L group, there is a complete lack of contrastive and entailment pairs (see Table \ref{tab:carto_distrib}). This issue should be fixed by the Cart-Stra-CL++ strategy, which makes sure to add examples from each class, in each difficulty batch. Cart-Stra-CL++ surpasses the Length-CL \cite{Nagatsuka-NGC-2023}, STS-CL, Cart-CL and Cart-CL++ strategies, as well as the vanilla Ro-BERT, confirming our intuition. Still, the \emph{overall performance} (Table \ref{tab:overallresults2}) on RoNLI does not exceed 80\% in terms of the micro $F_1$ (accuracy), which highlights both the need for the development of more robust and accurate models, and the research opportunity offered by our RoNLI dataset.

\vspace{-2.6mm}
\paragraph{Statistical testing.}
We performed a Cochran's Q statistical test to compare the Ro-BERT based on oversampling (micro $F_1= 0.73$, macro $F_1 = 0.56$) with the proposed Ro-BERT + Cart-Stra-CL++ (micro $F_1 = 0.75$, macro $F_1 = 0.59$). The test indicates that the proposed model is significantly better than the Ro-BERT based on oversampling, with a p-value of $0.001$. Note that the Cochran's Q test is applied on the contingency tables. Hence, the test rather indicates that the predictions are significantly different, without particularly using the reported micro / macro $F_1$ scores. 
We also performed the Mann-Whitney U test between Ro-BERT with oversampling and Ro-BERT+Cart-Stra-CL++. The Mann-Whitney U test further confirms that our method is significantly better than the baseline, with a p-value lower than $0.001$. 

\begin{table}[t]
\centering
\small
\setlength\tabcolsep{0.1cm}
\begin{tabular}{lcc}
\hline
\textbf{Method} & \textbf{Micro $\mathbf{F_1}$} &  \textbf{Macro $\mathbf{F_1}$} \\
\hline
BERT & {0.748} & 0.747 \\
\hline
{BERT + Length-CL} & {0.709} & 0.705 \\
BERT + Cart-Stra-CL++ & {\color{blue} \bf 0.756} & {\color{blue} \bf 0.755} \\
\hline
\end{tabular}
\vspace{-2mm}
\caption{Overall micro and macro $F_1$ scores of BERT, BERT+Length-CL and BERT+Cart-Stra-CL++ on the SciNLI \cite{sadat-caragea-2022-scinli} test set. The best results are shown in bold blue.}
\label{tab:results_scinli}
\vspace{-2mm}
\end{table}

\vspace{-2.6mm}
\paragraph{Generalization results.} To assess the generalization capacity of our novel learning method (Cart-Stra-CL++), we extend the evaluation to an additional dataset, namely SciNLI \cite{sadat-caragea-2022-scinli}. We compare three models on SciNLI: BERT \cite{Devlin-NAACL-2019}, BERT+Length-CL \cite{Nagatsuka-NGC-2023} and BERT+Cart-Stra-CL++ (ours). All BERT models use the English base cased version. 
Note that SciNLI is balanced, so oversampling is not needed for any of the methods. The results on SciNLI are shown in Table \ref{tab:results_scinli}. Our curriculum learning method achieves the best performance among the three methods, showing that Cart-Stra-CL++ generalizes to other datasets.

\section{Discussion}

A unique characteristic of RoNLI, by contrasting it with other existing NLI datasets, is that the hypotheses in RoNLI are sentences from Wikipedia which are not written by crowd-workers with the NLI task in mind, and these hypotheses are more diverse. In contrast, hypotheses in SNLI \cite{bowman-etal-2015-large} and MNLI \cite{N18-1101} are written by paid crowd-workers and lack diversity. For example, in SNLI, we observed many pairs such as those presented in Table \ref{tab:examples_snli} from Appendix \ref{sec:appendix}, which are less challenging for a model and are predicted as \emph{contradiction} by models without even looking at the first sentence (the premise). This issue was also observed by \citet{gururangan-etal-2018-annotation}.

Furthermore, we emphasize that Romanian is a unique Eastern Romance language, being part of a linguistic group that evolved from several dialects of Vulgar Latin which separated from the Western Romance languages between the 5th and the 8th centuries AD. Being surrounded by Slavic neighbors, it has a strong Slavic influence, which makes Romanian a unique language (among the Latin languages). Therefore, the results obtained by multilingual BERT in the zero-shot and fine-tuned settings (presented in Tables \ref{tab:results} and \ref{tab:overallresults2}) suggest that there are linguistic differences between our dataset for NLI in Romanian and other NLI-supported languages. This confirms the need for a language-specific dataset such as RoNLI, despite the presence of other Latin languages, such as French and Spanish, and even Slavic languages, such as Bulgarian and Russian, in XNLI. 

\section{Conclusion}

In this work, we introduced RoNLI, the first public corpus for Romanian natural language inference. We trained and evaluated several models to establish a set of competitive baselines for future research on our corpus. We also performed experiments to analyze the effect of spurious correlations, as well as the effect of harnessing data cartography to establish groups of useful samples or to develop curriculum learning strategies. Notably, we introduced a novel curriculum learning strategy based on data cartography and stratified sampling, boosting the overall micro and macro $F_1$ scores of Ro-BERT by $2\%$ and $3\%$, respectively. These improvements are statistically significant.
Nevertheless, our empirical results indicate that there is sufficient room for future research on the Romanian NLI task. We make our dataset and code available to the community to lay the grounds for future research on Romanian NLI. 

\section{Acknowledgments}

We thank reviewers for their constructive feedback, leading to significant improvements of our work.

\section{Limitations}

Our work is slightly limited by the number of samples with manual labels included in our corpus. This limitation is caused by the laborious annotation process involving human effort. However, we underline that RoNLI is not the only NLI corpus with training data based on automatic labels. For example, SciNLI \cite{sadat-caragea-2022-scinli} is created using a very similar process, based on automatically labeling English sentence pairs via linking phrases.

\section{Ethics Statement}

The manual labeling was carried out by volunteering students, who agreed to annotate the news articles in exchange for bonus points. Prior to the annotation, they also agreed to let us publish the labels along with the dataset. We would like to emphasize that the students understood that the annotation task is optional, and they could also get the extra bonus points from alternative tasks. Moreover, all students had the opportunity to get a full grade without the optional tasks. Hence, there was no obligation for any of the students to perform the annotations.

Our data is collected from Wikipedia, which resides in the public web domain. We note that the European regulations\footnote{\url{https://eur-lex.europa.eu/eli/dir/2019/790/oj}} allow researchers to use data in the public web domain for non-commercial research purposes. Thus, we release our data and code under the CC BY-NC-SA 4.0 license\footnote{\url{https://creativecommons.org/licenses/by-nc-sa/4.0/}}.

We acknowledge that some sentences could refer to certain public figures. To ensure the right to be forgotten, we will remove all references to a person, upon receiving removal requests via an email to any of the authors.

\bibliography{anthology,custom}

\appendix

\begin{table*}[th]
    \centering
    \small
    \begin{tabular}{lllc}
    \hline
    \textbf{Category} & \textbf{Romanian} & \textbf{English} & \textbf{Occurrences}\\ 
    \hline
    \multirow{19}{1em}{Contrastive} & Pe de alt\u{a} parte & On the other hand & 1981 \\
    & \^{I}n contrast & In contrast & 616\\ 
    & \^{I}n ciuda acestui fapt & In spite of this fact & 267\\
    & \^{I}n opozi\c{t}ie  & In opposition & 49\\
    & \^{I}n contradic\c{t}ie & In contradiction & 40\\
    & \^{I}n ciuda acestui lucru & In spite of this thing & 33\\
    & \^{I}n ciuda acestor fapte & In spite of these facts & 23 \\
    & \^{I}n ciuda acestor lucruri & In spite of these things & 11\\
    & \^{I}n mod contrar & In an opposite way & 11\\
    & Pe de cealalt\u{a} parte & On the other hand & 11\\
    & Cu toate acestea \^{i}ns\u{a} & Nevertheless & 9\\
    & Contrast\^{a}nd & In contrast & 5 \\
    & \^{I}n dezacord & In disagreement & 4\\
    & \^{I}n sens opus & In the opposite sense & 3\\
    & \^{I}n antiteza & In antithesis & 3\\
    & \^{I}n contradictoriu & Contradictory & 2\\
    & \^{I}ntr-un contrast & In a contrast & 2 \\
    & Contrar convingerilor & Contrary to the beliefs & 1\\
    & \^{I}n pofida acestor lucruri & Contrary to these things & 1\\
    \hline
    \end{tabular}   
    \caption{Original (Romanian) and translated (English) linking phrases and transition words used to retrieve contrastive sentence pairs, along with the number of retrieved sentences.}
    \label{tab:linking_contrastive}
\end{table*}

\begin{table*}[!ht]
    \centering
    \small
    \begin{tabular}{lllc}
    \hline
    \textbf{Category} & \textbf{Romanian} & \textbf{English} & \textbf{Occurrences}\\
    \hline
    \multirow{19}{1em}{Entailment} & Cu alte cuvinte & In other words & 553\\     
    & Adic\u{a} & Specifically &  	296 \\
    & \^{I}n esen\c{t}\u{a} & In essence & 155\\
    & Altfel spus & In other words & 149 \\
    & Asta \^{i}nseamn\u{a} c\u{a} & This means that & 92 \\
    & \^{I}n fond & In essence & 53\\
    & Sintetiz\^{a}nd & Synthesizing & 13\\
    & Rezum\^{a}nd & Summarizing & 12\\
    & \^{I}n rezumat & In summary & 10\\
    & \^{I}n termeni simpli & In simpler terms & 7\\
    & \^{I}n traducere liber\u{a} & In translation & 6\\
    & Mai pe scurt & In short & 6\\
    & \^{I}n al\c{t}i termeni & In other terms & 5\\
    & Simplific\^{a}nd & Simplifying & 5\\
    & Simplu spus & Simply said & 3\\
    & Mai concis & More concisely & 2\\
    & Pe larg & In broader terms & 5\\ 
    & \^{I}n termeni populari & In more popular terms & 1\\
    & \^{I}ntr-o alt\u{a} formulare & In another form & 1\\
    \hline
    \end{tabular}  
       \caption{Original (Romanian) and translated (English) linking phrases and transition words used to retrieve entailment sentence pairs, along with the number of retrieved sentences.}
           \label{tab:linking_entailment}
\end{table*}

\begin{table*}[!ht]
    \centering
    \small
    \begin{tabular}{lllc}
    \hline
    \textbf{Category} & \textbf{Romanian} & \textbf{English} & \textbf{Occurrences}\\
    \hline
    \multirow{24}{1em}{Causal}   & Astfel & Therefore & 16245\\ 
    & Prin urmare & As a consequence & 5202\\
    & Ca urmare & As an outcome & 4433\\
    & \^{I}n consecin\c{t}\u{a} & Consequently & 1010\\
    & A\c{s}adar & Hence & 948\\
    & Drept urmare & As a result & 601 \\
    & \^{I}n acest fel & In this manner & 574\\ 
    & Ca rezultat & As a result & 528\\
    & Din aceast\u{a} cauz\u{a} & Because of this & 520 \\
    & Astfel c\u{a} & Thus & 230\\
    & \^{I}n concluzie & In conclusion & 197 \\
    & Rezultatul este & The result is & 105 \\
    & \^{I}n rezultat & In result & 36\\
    & Din aceast\u{a} cauza & Because of this & 17\\
    & Concluzion\^{a}nd & Concluding & 14\\
    & Pentru a finaliza & To finalize & 7\\
    & Ca o consecin\c{t}\u{a} a acestui fapt & As a consequence of this & 4\\
    & \^{I}ntr-o concluzie & In a conclusion & 3 \\
    & Ceea ce a dus la & This lead to & 2\\
    & Duc\^{a}nd la & Leading to & 2\\
    & Conduc\^{a}nd la & Leading to & 1\\
    & Provoc\^{a}nd astfel & Thus causing & 1 \\
    & Se poate concluziona c\u{a} & It can be concluded that & 1 \\
    & \c{T}in\^{a}nd cont de acestea & Considering these& 1 \\
    \hline
    \end{tabular}  
    \caption{Original (Romanian) and translated (English) linking phrases and transition words used to retrieve reasoning sentence pairs, along with the number of retrieved sentences.}
        \label{tab:linking_causal}
\end{table*}

\section{Data Collection and Annotation}
\label{sec:appendix}

In this appendix, we present additional details about the data collection and annotation process.

\paragraph{Definition of sentence relationships.}
We use the same four relationship categories as \citet{sadat-caragea-2022-scinli}. These categories are defined as follows:
\begin{itemize}
    \item \textbf{Contrastive:} captures pairs of sentences where one statement either opposes, contrasts with, criticizes, or points out a limitation in relation to the other.
    \item \textbf{Entailment:} sentence pairs are structured such that the first sentence provides a foundation, cause, or precondition for the outcome articulated in the second sentence.
    \item \textbf{Reasoning:} represents pairs of sentences where one statement explains, refines, generalizes, or conveys a meaning akin to the other.
    \item \textbf{Neutral:} This group consists of sentence pairs that are semantically unrelated. For example, sentences that are targeting completely different subjects.
\end{itemize}
 
\paragraph{Linking phrases used for automatic labeling.} 
We next provide the specific linking phrases and transition words that are used to retrieve the sentence pairs from Wikipedia. These are listed in Table \ref{tab:linking_contrastive} for the contrastive pairs, Table \ref{tab:linking_entailment} for the entailment pairs, and Table \ref{tab:linking_causal} for the causal pairs. There are no linking phrases to be listed for the neutral sentences. For each linking phrase, we report the number of extracted sentences.

We underline that after automatically assigning the labels, the linking phrases are removed from the sentences, making the NLI task much harder. Since we use the linking phrases to automatically label the training samples, leaving these phrases in place would make the task too simple for automated models. We argue that models should not only rely on obvious cues (such as the linking phrases) to reach human-level capabilities in NLI. We thus believe that the designed task is more suitable for the desired end goal. Moreover, the performance levels of the tested models are well above random chance, indicating that there are sufficient clues left for models to learn. Nevertheless, we perform an experiment where the linking phrases are included and the performance of Ro-BERT increases by considerable margins, reaching a micro $F_1$ of  $0.81$ and macro $F_1$ of $0.68$ on the test set. The error rate of this model can be explained by the fact that the test set is labeled manually, so overfitting to the linking phrases is not necessarily the best solution.

\begin{table*}
\setlength\dashlinedash{0.5pt}
\setlength\dashlinegap{1.2pt}
\setlength\tabcolsep{4.7pt}
\centering
\small
\begin{tabular}{p{40.2em} p{8em}}
\hline
{\textbf{Premise}} & {\textbf{Hypothesis}}\\
\hline
A man in a blue shirt is performing a skateboarding trick near stairs while two other men watch. & Nobody has a shirt. \\
\hdashline 
Two young children, one wearing a red striped shirt, are looking in through the window while an adult in a pink shirt watches from behind. & Nobody has a shirt.\\
\hdashline
A boy in a yellow t-shirt and pink sweater talks on a cellphone while riding a horse through a crowd of people who are looking on. & Nobody has a shirt.\\
\hline
\end{tabular}
\vspace{-2mm}
\caption{Examples of sentence pairs from SNLI that exhibit annotation artifacts \cite{gururangan-etal-2018-annotation}.}
\label{tab:examples_snli}
\end{table*}

\paragraph{Annotator selection.}
We offered the possibility of annotating sentence pairs to students enrolled at the AI/NLP Master programs from the University of Bucharest, in exchange for bonus points awarded for their final grades. We specified the exact benefits awarded for annotating between 6K and 7K sentence pairs, so the students knew exactly what to expect. The students can have either computer science or linguistics background. We enrolled the first three Master students who volunteered for this optional assignment. All annotators are native Romanian speakers. Aside from the instructions presented below, all students attended a lecture in which the NLI and NLU tasks were extensively discussed.

\paragraph{Instructions to annotators.}
The instructions given to the annotators, translated from Romanian to English, are as follows:

\emph{Natural language inference (NLI) is the task of recognizing the relationship in sentence pairs. In this labeling task, you will be presented with sentence pairs of the form (Sentence A, Sentence B), without any additional context. Your task is to determine the relationship between sentences A and B, choosing one of the following four options:
\begin{itemize}
    \item Contrastive: Select this category if one sentence presents a viewpoint or fact that is different from the other. Sentence B does not have to directly oppose Sentence A. Hence, this category includes comparisons, criticisms, or pointing out a limitation or unique aspect in one sentence about the content of the other.
    \item Reasoning: Choose this option if Sentence A provides a basis or rationale that leads to or explains the information in Sentence B. Look for a logical sequence where the first sentence sets up a foundation that the second sentence builds upon or concludes from.
    \item Entailment: Use this category if one sentence expands on or specifies the information given in the other, essentially providing more details or a specific instance of the general idea of the premise.
    \item Neutral: Opt for this category if the two sentences are unrelated, with each standing independently without referring to, supporting, or elaborating on the other. A sentence pair that does not fit in the other categories should be labeled as neutral. 
\end{itemize}}
The above instructions were followed by one pair of sentences from each category, to exemplify the four categories. The provided examples were manually labeled by the authors, to avoid any potential mistakes resulting from the automatic labeling process.

\begin{table*}
\setlength\dashlinedash{0.5pt}
\setlength\dashlinegap{1.2pt}
\setlength\tabcolsep{4.7pt}
\centering
\small
\begin{tabular}{p{10em} p{10em} p{10em} p{10em} p{5.2em}}
\hline
\multicolumn{2}{c}{\textbf{Romanian}} & \multicolumn{2}{c}{\textbf{English}} & \multirow{2}{*}{\textbf{Class}} \\
{\textbf{Premise}} & {\textbf{Hypothesis}} & {\textbf{Premise}} & {\textbf{Hypothesis}} & \\
\hline
Sub umbra soacrei ei, Ulrica nu a fost niciodat\u{a} fericit\u{a} sau cel pu\c{t}in via\c{t}a ei de la curte nu a fost fericit\u{a}. & S-a spus c\u{a} via\c{t}a ei privat\u{a} cu regele \c{s}i copiii ei a fost una foarte fericit\u{a}. & Under the shadow of her mother-in-law, Ulrica was never happy, or at least her life at court was not happy.& It was said that her private life with the king and her children was a very happy one.  & Contrastive \\
\hdashline 
\^{I}n teoria jocurilor, un joc cu sum\u{a} zero sau nul\u{a} descrie o situa\c{t}ie \^{i}n care c\^{a}\c{s}tigul unui participant este perfect echilibrat pierderea unui alt participant.  &  Orice situa\c{t}ie care func\c{t}ioneaz\u{a} ca un joc cu sum\u{a} zero presupune c\u{a} orice c\^{a}\c{s}tig al unui participant necesit\u{a} o pierdere egal\u{a} a altui participant. & In game theory, a zero sum or null sum game describes a situation in which the gain of a participant is balanced by the loss of another participant. & Any situation that works like a zero sum game assumes that any gain of a participant will require the equal loss of another participant. & Entailment\\
\hdashline
Cum pie\c{t}ele bursiere au sc\u{a}zut \^{i}n septembrie 2008, fondul a putut s\u{a} cumpere multe ac\c{t}iuni la pre\c{t}uri sc\u{a}zute. & Pierderile suportate de turbulen\c{t}ele de pe pie\c{t}e au fost recuperate p\^{a}n\u{a} \^{i}n noiembrie 2009. & As the stock market fell in September 2008, the fund was able to buy more stocks at lower prices & The losses incurred by the market turbulence were recovered by November 2009. & Reasoning\\
\hdashline
Conform Catalogue of Life specia ``Drosophila gibberosa'' nu are subspecii cunoscute. & Interesul s\u{a}u \^{i}n domeniul meteorologiei l-a determinat s\u{a} se ocupe de avia\c{t}ie. & According to the Catalog of Life, the species ``Drosophila gibberosa'' has no known subspecies. & His interest in meteorology led him to pursue aviation. & Neutral\\
\hline
\end{tabular}
\vspace{-2mm}
\caption{Original and translated examples of sentence pairs extracted from the Romanian Wikipedia, which are automatically labeled via linking phrases and transition words. We show one example per class.}
\label{tab:examples}
\end{table*}

\begin{table}[t]
\centering
\small
\setlength\tabcolsep{0.15cm}
\begin{tabular}{lccc}
\hline
\textbf{Group} & \textbf{E2L} & \textbf{A} & \textbf{H2L} \\
\hline
Fleiss Kappa & 0.75 & 0.72 & 0.69\\
\hline
\end{tabular}
\vspace{-2mm}
\caption{Inter-rater agreements for easy-to-learn (E2L), ambiguous (A) and hard-to-learn (H2L) samples from the validation set.}
\vspace{-3mm}
\label{tab:carto_agree}
\end{table}

\begin{table*}[t]
\setlength\dashlinedash{0.5pt}
\setlength\dashlinegap{1.2pt}
\setlength\tabcolsep{4.7pt}
\centering
\small
\begin{tabular}{p{12.3em} p{7.7em} p{12.3em} p{7.7em} p{5.2em}}
\hline
\multicolumn{2}{c}{\textbf{Romanian}} & \multicolumn{2}{c}{\textbf{English}} & \textbf{Correct Class /} \\
{\textbf{Premise}} & {\textbf{Hypothesis}} & {\textbf{Premise}} & {\textbf{Hypothesis}} & \textbf{Predicted Class}\\
\hline
Dac\u{a} nu autorit\u{a}\c{t}ile sunt cele care s\u{a} decid\u{a} doctrina, \c{s}i dac\u{a} argumentele lui Martin Luther pentru preo\c{t}ia tuturor credincio\c{s}ilor sunt duse la extrem, caz \^{i}n care Biserica ar fi condus\u{a} de cei ale\c{s}i, atunci a avea monarhul \^{i}n fruntea Bisericii ar fi intolerabil. & Dac\u{a} monarhul e numit de Dumnezeu s\u{a} fie \^{i}n fruntea bisericii, atunci e intolerabil ca parohiile locale s\u{a} \^{i}\c{s}i decid\u{a} singure doctrina. & If authorities are not the ones to decide the doctrine, and if the arguments of Martin Luther for the priesthood of all the believers are taken to the extreme, in which case the Church would be led by the chosen, then to have the monarch as the head of the Church would be intolerable. & If the monarch is named by God to be the head of the Church, then it is intolerable for local churches to decide their own doctrine. & Contrastive / Reasoning \\
\hdashline
Dar Biserica \^{i}nva\c{t}\u{a} c\u{a} realitatea \c{s}i eficacitatea Sfintelor Taine ale Bisericii realizate de preo\c{t}i nu depind de virtu\c{t}ile lor personale, ci de prezen\c{t}a \c{s}i lucrarea lui Iisus Hristos, Care ac\c{t}ioneaz\u{a} \^{i}n Biserica Sa prin Sf\^{a}ntul Duh. & Preo\c{t}ia este un dar al bisericii - trupul lui Hristos \^{i}n istorie -, \^{i}n slujba comunit\u{a}\c{t}ii ecleziale, iar nu un dar personal al celui care devine preot. & But the Church teaches that the efficacy of the Holy Teachings of the Church preached by priests do not depend on their personal virtues, but on the presence and work of Christ, Who acts in His Church through the Holy Spirit. & Priesthood is a gift of the Church, body of Christ in History - in the work of the ecclesiastical community, not a gift to the one who becomes a priest & Entailment / Reasoning\\
\hline
\end{tabular}
\vspace{-2mm}
\caption{Original and translated examples of sentence pairs that are misclassified by our best performing model (Ro-BERT + Cart-Stra-CL++).}
\label{tab:examples_misclassified}
\end{table*}

\paragraph{Automatic vs.~manual annotations.} In the main paper, we used Cohen's Kappa in order to determine the agreement between automatic and manual annotations. As an alternative way to estimate the alignment between automatic and manual annotations, we compute the micro and macro $F_1$ scores using automatic labels as predictions and manual labels as ground-truth. The resulting micro $F_1$ is $0.83$ and the macro $F_1$ is $0.69$. Notice that these scores are much higher than the machine learning models reported in Table \ref{tab:overallresults2}, confirming that the NLI task on our dataset is not yet saturated.

In Table \ref{tab:examples}, we show one randomly selected sentence pair per class. The examples are taken from the training set, being labeled automatically. We observe that the assigned labels are correct.

To determine if the H2L samples have a higher percentage of noisy labels, we applied data cartography to the validation set. After training Ro-BERT on the validation set to find the E2L, A, and H2L samples, we were able to compute the inter-rater agreement for each subgroup. We obtained the  Kappa coefficients reported in Table \ref{tab:carto_agree}. The reported inter-rater agreements confirm that the H2L group contains more examples with potentially wrong labels.

\section{Hyperparameter Tuning}

The hyperparameters of all models are determined via grid search. We used the following intervals for the various hyperparameters:
\begin{itemize}
    \item Learning rates between $10^{-2}$ and $10^{-6}$.
    \item Mini-batch sizes between $8$ and $256$ (using only powers of two as options).
    \item Number of hidden units for the classification head in the set $\{128, 256, 512, 768\}$.
    \item Dropout rates between $0.1$ and $0.5$.
    \item Values for the regularization hyperparameter $C$ of the SVM and Softmax models  between $0.1$ and $1000$.
    \item SVM kernel options between \emph{linear} and \emph{RBF}.
    \item Tolerance (for optimization) between $10^{-2}$ and $10^{-6}$.
\end{itemize}

For SVM, we obtain optimal validation results with $C=0.5$, the maximum number of iterations set to $2500$, and the tolerance level equal to $10^{-5}$. For Softmax, we reach the best validation results with $l_2$ regularization with $C=1$, and the tolerance level equal to $10^{-3}$.

The RoGPT2 model employs the AdamW optimizer \cite{loshchilov2019decoupled} with a learning rate of $10^{-3}$, on mini-batches of 64 samples. The model is trained for at most 10 epochs, with early stopping. The fine-tuned multilingual BERT is optimized with AdamW for 10 epochs on mini-batches of 256 samples, using a learning rate of $10^{-3}$. As the multilingual BERT baseline, the models based on Ro-BERT employ the AdamW optimizer with a learning rate of $10^{-3}$, on mini-batches of 256 samples. The models are trained for at most 10 epochs, with early stopping.

All other hyperparameters are set to their default values. Please note that we release the code to reproduce all baselines, along with the RoNLI corpus\footnote{\url{https://github.com/Eduard6421/RONLI}}.

\section{Error Analysis and Task Complexity} 

To find interesting language-specific phenomena in Romanian NLI, we perform an error analysis of the best scoring model (Ro-BERT + Cart-Stra-CL++). We present two examples mislabeled by our best model in Table \ref{tab:examples_misclassified}. The first example showcases a lack of understanding of Romanian sentence structure and formulation (sentences are mislabeled as reasoning instead of contrastive). The sentences exemplify complex Romanian phrasing structures with multiple clauses and conditional phrases. Such complexity can lead to classification errors, if the classifier is not apt at parsing and understanding the nuances in Romanian sentence construction. In the second example, both sentences are structurally complex, with multiple subordinate clauses. They share thematic elements related to Christianity, which may contribute to their classification error due to overlapping religious concepts. The error analysis reveals some interesting findings. For example, Romanian tends to have a complex sentence structure, which often confuses NLI models. 

We note that our assertions are supported by the study of \citet{Dobrovie-DGM-1994}. We summarize the main aspects that influence the complexity of the Romanian language below:
\begin{itemize}
    \item The clitic system of Romanian exhibits a level of complexity not found in other Romance languages, characterized by the presence of not only pronominal clitics, but also adverbial clitics and cliticized conjunctions. Significantly, Romanian includes auxiliary elements that serve as verbal clitics, distinguishing its syntactic structure further.
    \item The constituent structure of Romanian clauses, with particular emphasis on subjunctive clauses, displays marked distinctions from those observed in other Romance languages, as extensively discussed in Chapter 3 of \cite{Dobrovie-DGM-1994}. This divergence underscores the unique syntactic configurations inherent to Romanian.
    \item Romanian demonstrates a predilection for employing subjunctive constructions in contexts where other Romance languages would typically resort to infinitive forms. This syntactic preference highlights a notable deviation in structural utilization across Romance languages.
    \item The ambiguous nature of the particle `a' holds substantial comparative significance. Such ambiguous particles are absent in other Romance languages and English, yet they are prevalent in Verb-Subject-Object languages, including Welsh. This phenomenon presents an intriguing area of study from a comparative linguistic perspective.
\end{itemize}

\end{document}